\documentclass[11pt]{article}

\usepackage[final]{acl}
\usepackage{times}
\usepackage{latexsym}

\usepackage[T1]{fontenc}

\usepackage[utf8]{inputenc}

\usepackage{microtype}

\usepackage{inconsolata}

\usepackage{graphicx}
\usepackage{mathtools}
\usepackage{amsmath}
\usepackage{amssymb}
\usepackage{bm} 
\usepackage{listings}
\usepackage{xcolor}
\usepackage{subfig}
\usepackage{tikz}
\usetikzlibrary{arrows.meta, positioning}

\lstdefinelanguage{Cypher}{
  morekeywords={MATCH,RETURN,CASE,WHEN,THEN,ELSE,AS},
  sensitive=true,
  morecomment=[l]{//},
  morestring=[b]",
}

\lstdefinestyle{cypherStyle}{
  language=Cypher,
  basicstyle=\ttfamily\small,
  keywordstyle=\color{blue}\bfseries,
  stringstyle=\color{red},
  commentstyle=\color{gray},
  backgroundcolor=\color{gray!10},
  frame=single,
  breaklines=true,
  showstringspaces=false,
  columns=flexible
}
\usepackage[most]{tcolorbox}
\usepackage{amsmath}
\usepackage{amssymb}
\title{Decoding Plastic Toxicity: An Intelligent Framework for Conflict-Aware Relational Metapath Extraction from Scientific Abstracts}

\author{Sudeshna Jana, Manjira Sinha \and Tirthankar Dasgupta \\
      TCS Research\\ India\\
      \texttt{(sudeshna.jana, sinha.manjira, dasgupta.tirthankar)@tcs.com}}

\begin{document}
\maketitle
\begin{abstract}
The widespread use of plastics and their persistence in the environment have led to the accumulation of micro- and nano-plastics across air, water, and soil — posing serious health risks including respiratory, gastrointestinal, and neurological disorders. We propose a novel framework that leverages large language models to extract relational metapaths — multi-hop semantic chains linking pollutant sources to health impacts — from scientific abstracts. Our system identifies and connects entities across diverse contexts to construct structured relational metapaths, which are aggregated into a \textit{Toxicity Trajectory Graph} that traces pollutant propagation through exposure routes and biological systems. Moreover, to ensure consistency and reliability, we incorporate a dynamic evidence reconciliation module that resolves semantic conflicts arising from evolving or contradictory research findings. Our approach demonstrates strong performance in extracting reliable, high-utility relational knowledge from noisy scientific text and offers a scalable solution for mining complex cause-effect structures in domain-specific corpora.
\end{abstract}

\section{Introduction}

The extensive use of plastics across industries such as manufacturing, medicine, textiles, packaging, and cosmetics has led to the present era is often referred as the ``Plastic Age'' \cite{blocker2020living}. Due to their low cost, durability, and versatility, global plastic production has surged from 2.3 million tons in 1950 to 450 million tons in 2019, with projections to double by 2050 \cite{dokl2024global}. Meanwhile, annual plastic waste generation exceeds 350 million tons \cite{owid-plastic-pollution}, of which only 9\% is recycled, while the remainder is incinerated, landfilled, or mismanaged, contributing to environmental contamination. These discarded plastics degrade into micro- and nano-plastics (MNPs) through physical, chemical, and biological processes \cite{thompson2004lost, cai2021analysis}.

Consequently, plastic pollution is now pervasive across global ecosystems, posing severe risks to biodiversity and human health \cite{macleod2021global}. Due to their tiny size and persistence —MNPs disperse widely through air, soil, and water, entering food chains and eventually the human body. Exposure occurs mainly through ingestion, inhalation, and, to a lesser extent, dermal contact, leading to serious health concerns.

In recent years, researchers have increasingly focused on various plastic pollutants, a significant amount of research has been conducted on their occurrence in the environment and impacts on human health. \cite{gasperi2018microplastics, amato2020emerging, riaz2024breathing, huang2022detection} Studies showed the presence of MNPs in respiratory systems and increase susceptibility to lung disorders including chronic obstructive pulmonary disease, fibrosis, dyspnea, asthma, and the formation of frosted glass nodule. Smokers were found to have elevated levels of polyurethane and silicone particles in their lower respiratory tract \cite{lu2023new}. Occupational exposure varies by environment: outdoor workers (e.g., couriers) exhibited polycarbonate and PVC in sputum, while indoor workers (e.g., office staff) had PVC and Polyamide as dominant MNPs \cite{jiang2022exposure}. Several studies \cite{al2023microplastics, de2020microplastics, rainieri2019microplastics, cverenkarova2021microplastics} have investigated human MNPs exposure through food sources such as seafood, food additives, drinks, and plastic packaging, highlighting the risks of chronic biological effects and potential health hazards, including gastrointestinal, immune, reproductive, and respiratory disorders, as well as cancer and chromosomal anomalies. Though the skin provides partial protection, recent studies \cite{aristizabal2024microplastics, wang2023microplastics, menichetti2024penetration, ouyang2025microplastics} indicate entry via sweat glands, lesions, or follicles, leading to inflammation and accelerated aging. In addition to the studies mentioned, numerous other research reports provide evidence of how MNPs have become ubiquitous in human body, further highlighting their potential threat to human health.

\begin{figure}[ht]
\includegraphics[width=\linewidth]{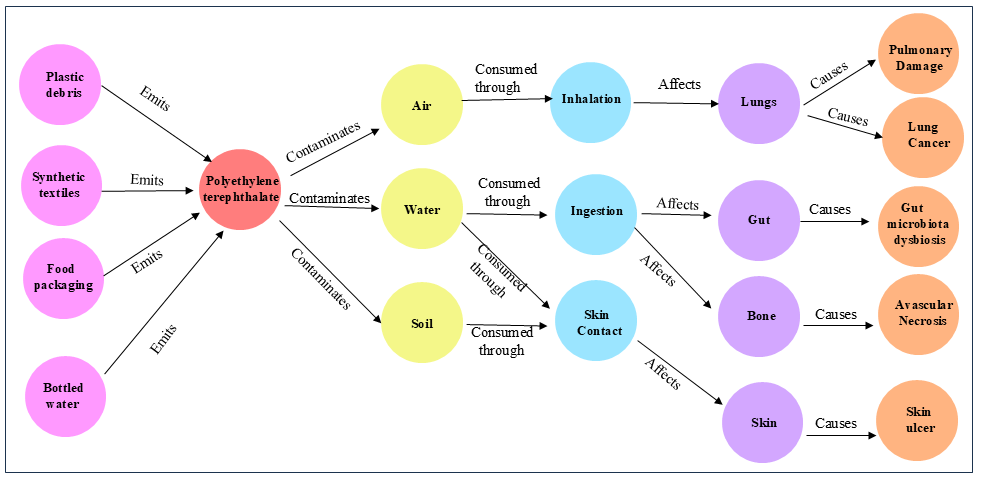}
\centering
\caption{Toxicity Trajectory for `Polyethylene terephthalate', illustrating sequential relational transitions among different node types.}
\label{sample_path}
\end{figure}

However, the rapid growth of research on plastic pollution has made it increasingly difficult for researchers and public health professionals to stay up-to-date with emerging findings and systematically identify harmful pollutants, their sources, exposure pathways, and associated health outcomes. To address these challenges, we analyze a large corpus of PubMed abstracts to construct a structured scientific knowledge graph, termed \textit{Toxicity Trajectory Graph}, which encodes relational toxicity pathways by integrating fragmented information on pollutant sources, exposure routes, affected biological systems, and health outcomes. Figure~\ref{sample_path} shows a representative sub-trajectory for `Polyethylene terephthalate'. The major contributions of this paper are summarized as follows:

\begin{itemize}
    \item We propose a large language model (LLM) based framework for extracting relational metapaths that represent toxicity pathways of plastic pollutants, leveraging a large corpus of PubMed abstracts.
    
    \item A consistency evaluation module is introduced to detect and resolve contradictions arising from evolving scientific evidence, ensuring the coherence and reliability of the extracted relational metapaths.
    
    \item Finally, the \textit{Toxicity Trajectory Graph} is constructed to capture multi-hop relational pathways across a large set of plastic pollutants, linking sources, exposure routes, biological mechanisms, and health outcomes affecting diverse population groups.
\end{itemize}


\section{Dataset}\label{data}
We collected abstracts from 5,282 articles published between 2012 and the present, sourced from the publicly available PubMed archive\footnote{\url{https://pubmed.ncbi.nlm.nih.gov/}}. To retrieve relevant literature, we used the \textit{Metapub} library and constructed queries using various combinations of the following keywords: `plastic', `microplastic', `nanoplastic', `pollution', `human', `health', `effects' and `disease'. Each abstract was assigned a unique PubMed ID (PMID), and metadata such as author information and publication year were recorded to support temporal analyses of how pollutant–health relationships have evolved over time. The average abstract length is approximately 300 tokens. All abstracts were preprocessed to remove extraneous content, including HTML tags, URLs, and other noisy artifacts. These PubMed abstracts offer rich insights into various plastic pollutants, including their sources, exposure routes, health effects, and affected populations. Figure~\ref{sample_abstract} in Appendix \ref{appendix: sample_abstract} shows a sample abstract with highlighted entity types, illustrating the diversity of information captured.

\section{Proposed Framework for Constructing Toxicity Trajectory Graph}\label{Method}
\begin{figure*}
\begin{center}
\includegraphics[width=.9\textwidth]{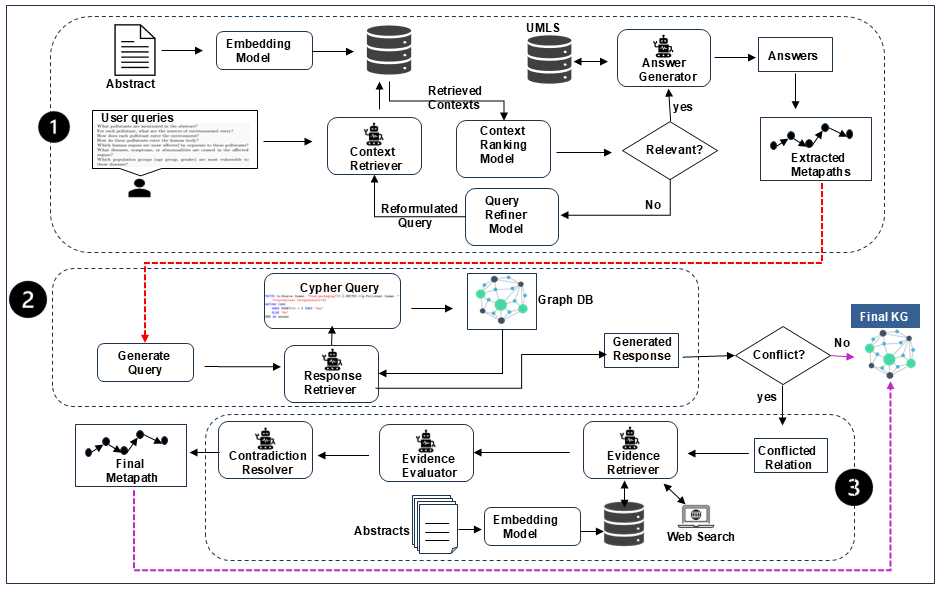}   
\end{center}
\caption{Schematic overview of our framework. The figure highlights the pipeline for relational metapath extraction ( 1 in top), relational consistency evaluation (2 in middle) and relational disagreement resolution system if any conflict occurs(3 in bottom).
 }
\label{framework}
\end{figure*}

Our proposed framework, illustrated in Figure~\ref{framework}, comprises three main components: (1) extraction of relational metapaths from PubMed abstracts, (2) evaluation of relational consistency to detect contradictions among metapaths involving the same entity pairs, and (3) resolution of contradictions to construct a coherent and reliable knowledge graph, \textit{Toxicity Trajectory Graph}.

\subsection{Relational Metapaths Extraction}
\label{metapath_extraction}
The first stage of our framework involves extracting relational metapaths by aggregating relational triples derived from PubMed abstracts, with a particular focus on capturing toxicity pathways related to various pollutants. Formally, a relational metapath $M$ is defined as a path represented in the form $E_1\xrightarrow{\text{R1}}E_2\xrightarrow{\text{R2}}E_3\ldots E_n\xrightarrow{\text{Rn}}E_{n+1}$, where $E_1$, $E_2$,...., $E_{n+1}$ are the entities or node and $R1$, $R2$,...., $Rn$ are the relations between those entities. The composite relation  $R = R_1 \circ R_2 \circ R_3 \circ \dots \circ R_n$ describes the connection between node $E1$ and $E_{n+1}$. 
In particular, our metapaths consist of the following six type of nodes:
\begin{itemize}
    \item \textbf{Pollutant (P):} Specific plastic pollutants mentioned in the abstract.
    \item \textbf{Source (S):} Environmental origins (e.g., microbeads, tire wear, synthetic textiles).
    \item \textbf{Medium (M):} Environmental carriers such as air, water, or soil.
    \item \textbf{Exposure Route (R):} Human exposure pathways, such as ingestion, inhalation, dermal contact.
    \item \textbf{Organ (O):} Affected biological systems or organs.
    \item \textbf{Disease (D):} Associated health outcomes, including demographic markers (e.g., age, gender).
\end{itemize}


Furthermore, the relationships among these nodes are defined as follows:
S$\xrightarrow{\text{emits}}$ P  $\xrightarrow{\text{contaminates}}$ M $\xrightarrow{\text{consumedthrough}}$ R $\xrightarrow{\text{affects}}$ O $\xrightarrow{\text{causes}}$ D. In addition to positive relations, some abstracts explicitly mention negative associations between entity pairs. We identify and incorporate three key types of negated relations into our pipeline: S $\xrightarrow{\text{not\_emit}}$ P, P $\xrightarrow{\text{not\_affect}}$ O, and P $\xrightarrow{\text{not\_cause}}$ D. Integrating these negations enhances the informativeness and expressiveness of the constructed knowledge graph. The following section outlines the pipeline for relational metapath extraction from abstract.

\subsubsection{Context Retriever}
The pipeline begins with a predefined set of queries $Q$ (see Appendix~\ref{User_Defined_Queries}) aimed at extracting relational triples from PubMed abstracts. Each abstract is segmented into context-aware textual chunks, which are then embedded using the \textit{all-MiniLM-L6-v2} Sentence Transformer~\cite{reimers-2019-sentence-bert}. The resulting embeddings are indexed using a chroma vector store to support efficient semantic retrieval. For each query $q \in Q$, a Context Retriever agent built upon \textit{Llama-3.1-8B-Instruct} language model, interprets the intent and retrieves the top-$k$ most relevant context chunks $C_q = {c^q_1, c^q_2, \ldots, c^q_k}$ via vector similarity search.

\subsubsection{Context Ranker}
Afterward, a graph-based context ranking model is implemented to rank the retrieved contexts $C_q$ by their semantic relevance to a query $q$. A graph $G = (V, E)$ is constructed with nodes $V = {v_0, v_1, \dots, v_k}$, where $v_0$ represents the query $q$ and $v_i$ corresponds to context $c_i \in C_q$. Edges in $E$ capture both query-to-context relevance and inter-context semantic similarity. Initial node embeddings $h_i$ for contexts $c_i$ and $h_q$ for question $q$ are generated using a BERT-based encoder:
\begin{align*}
    h_i &= BERT\ \left(\left[CLS\right]q\left[SEP\right]c_i\right)_{\left[CLS\right]} \\
    h_q &= BERT\ \left(\left[CLS\right]q\left[SEP\right]\right)_{\left[CLS\right]}
\end{align*}

Each $h_i$ is further concatenated with two auxiliary features: percentage of shared tokens between $q$ and $c_i$, and the count of overlapping named entities.
We employ a two-layer graph attention network (GAT) to iteratively update node representations by aggregating information from neighboring nodes via attention mechanisms.\cite{velivckovic2017graph} Specifically, each node’s representation $h_i$ is updated to a new representation $h_i^\prime$ using attention-weighted aggregation as follows:
\begin{align*}
h_i^\prime &= \sigma\!\left(\sum_{j\in\mathcal{N}(i)} \alpha_{ij}\cdot W h_{j}\right) \\
\alpha_{ij} &= \mathrm{softmax}_j\!\left(\mathrm{LeakyReLU}\!\left({a}^\top \left[\,{W}{h}_i \,\middle|\, {W}{h}_j \right]\right)\right)
\end{align*}

 ,where $\mathcal{N}(i)$ denotes the set of neighboring nodes of node $i$, $W$ is a learnable weight matrix and ${a}$ is the learnable attention vector. $\alpha_{ij}$ represents the attention weights from node $i$ to node $j$, and $\sigma$ is the activation function. To train the GAT in a self-supervised manner, we define the following contrastive loss function based on Information Noise-Contrastive Estimation (\textit{InfoNCE}). 
\begin{equation*}
    \mathcal{L}_{\mathrm{InfoNCE}} = -\log\frac{\exp{\left(\ \mathrm{sim}\left(h_q^\prime, h_{{c}^+}^\prime\right)/\tau\right)}}{\sum_{j=1}^{k}\exp{\left(\mathrm{sim}\left(h_{q}^\prime,{h}_{{c}_{j}}^\prime\right)/\tau\right)}}
\end{equation*}
Here, $\mathrm{sim}(\cdot, \cdot)$ denotes cosine similarity, $\tau$ is a temperature hyperparameter, $h_q^\prime$ and $h_{c_j}^\prime$ are the GAT-encoded embeddings of the query and context nodes, respectively. For each query $q$, a pseudo-positive context $c^+ \in C_q$ is selected based on the highest cosine similarity between the BERT embeddings of $q$ and $c_j$.
The model is trained to align the query embedding with the positive context while pushing it away from less relevant ones, thereby learning to distinguish semantically aligned context-question pairs. During inference, cosine similarity scores are computed between the final query and each context embedding; contexts scoring above a predefined threshold $0.9$ are marked as relevant. The top-$n$ relevant contexts are then forwarded to the Answer Generation module. If no context meets the threshold, a fallback mechanism is triggered, optionally invoking the query refinement model.

\subsubsection{Query Refiner:}
If all retrieved contexts are deemed irrelevant, an Entity-Guided Query Refinement module is triggered to generate a more effective query. First, named entities are extracted from the original query $q$ and the initially retrieved but irrelevant contexts $C_q = \{c_1, c_2, \ldots, c_k\}$ using the SpaCy NER toolkit\cite{}. Let $\epsilon_q$ denote entities from the query and $\epsilon_{ctx} = \bigcup_j \epsilon_{c_j}$ the union of entities from all retrieved contexts. To identify missing concepts, we compute: $\epsilon_{miss} = (\epsilon_q \cup \epsilon_{related}) \setminus \epsilon_{ctx}$
where $\epsilon_{related}$ includes semantically enriched concepts (e.g., synonyms, co-mentioned terms) retrieved from the neighborhood of $\epsilon_q$ in an external knowledge graph such as Wikidata\cite{}. Since critical concepts are not always explicitly mentioned in the original query, semantic enrichment helps uncover related but implicit entities that may improve context retrieval upon query reformulation.

A prompt is constructed using the original query $q$, irrelevant contexts $C_q$, and missing entities $\epsilon_{\text{miss}}$, which serves as the policy input to train LLaMA-3.1–8B-Instruct model via reinforcement learning. The model generates a refined query $q^\prime$, which is used to retrieve updated contexts $C_{q^\prime}$. The quality of refinement is evaluated using a reward based on the recovery of missing entities:

A prompt is constructed using the original query $q$, irrelevant contexts $C_q$, and missing entities $\epsilon_{\text{miss}}$, which serves as the policy input to train LLaMA-3.1–8B-Instruct model via reinforcement learning. The model generates a refined query $q^\prime$, and the retrieved contexts $C_{q^\prime}$ are evaluated by the fraction of recovered missing entities. Optimization is performed via REINFORCE using the following reward $r$ and loss functions $\mathcal{L}_\mathrm{R}$:
\begin{align*}
r &= \frac{|\epsilon_{\text{miss}} \cap \epsilon_{ctx^\prime}|}{|\epsilon_{\text{miss}}|}, \\
\mathcal{L}_\mathrm{R} &= -\log \pi\theta(q^\prime \mid s) \cdot (r - b)
\end{align*}

where $\pi_\theta$ denotes the LLM, $s$ is the input prompt, $r$ is the reward, and $b$ is a baseline for variance reduction. The refined contexts $C_{q^\prime}$ are subsequently used by downstream modules, including the Context Ranker and Metapath Generator.

\subsubsection{Metapath Generator:} 
Once relevant contexts are retrieved, an answer generation module based on LLaMA-3.1–8B-Instruct synthesizes responses to each query $q$ by integrating the retrieved information. However, generated answers often suffer from terminological inconsistency due to the heterogeneous vocabulary across biomedical literature. For instance, `Polyvinyl Chloride' may appear as `PVC’, `vinyl', or `Polychloroethylene’; likewise, `liver cancer' may be described as `hepatoma’ or `malignant neoplasm of liver’. To address this, we integrate the UMLS Metathesaurus API \cite{bodenreider2004unified} into the generation pipeline to normalize biomedical entities, ensuring terminological standardization and semantic consistency.

For each abstract, the generated answers are parsed into relational triplets grounded in predefined relation types. These triplets are then grouped by entity types -- for example, (Vehicle, emits, CO2) maps to S $\xrightarrow{\text{emits}}$ P. We construct a multilayer heterogeneous graph using NetworkX, where nodes represent entities, edges encode relations, and layers correspond to entity types. Valid relational metapaths are extracted via breadth-first search (BFS), beginning from the first-layer entity type and terminating at a designated endpoint. \cite{then2014more} The resulting set of relational metapaths is further evaluated for temporal alignment and factual correctness in subsequent modules.

\subsection{Relational Consistency Evaluation}
In this step, we employ a graph-based agentic framework that assesses the compatibility of new metapaths with previously observed relationships to ensure temporal coherence and factual consistency of the extracted relational metapaths. For each extracted metapath $m \in M$, we generate a set of closed-ended verification queries $Q_m$ by verbalizing its sequential edges into natural language, which are then translated into Cypher queries. An illustrative example is provided in Appendix~\ref{appendix_cypherquery}. These queries are executed by a graph-based LLM agent built upon LLaMA-3.1–8B-Instruct, over a Neo4j knowledge graph containing all prior metapaths. If a query matches existing relations, a contextual response is synthesized, otherwise, the relation is marked as novel. In case of conflict, a disagreement resolution module is invoked to assess its validity. If consistent, the metapath is integrated into the final knowledge graph. 



\subsection{Relational Disagreement Resolution}
To enhance the reliability and factual consistency of the constructed knowledge graph, it is crucial to resolve conflicting relational metapaths extracted from different abstracts. For example, one source may state that microplastics contribute to Crohn’s disease, while another denies any causal link—yielding contradictory relations such as: \textit{Microplastic} $\xrightarrow{\text{cause}}$ \textit{Crohn's disease} and \textit{Microplastic} $\xrightarrow{\text{not\_cause}}$ \textit{Crohn's disease}. To address such inconsistencies, we employ an agentic AI system comprising three collaborative agents built upon LLaMA-3.1–8B-Instruct, that assess contextual evidence, infer factual stance, and reconcile contradictions. The system resolves inconsistencies based on confidence scores and source credibility, ensuring the knowledge graph remains coherent while integrating new information.

\subsubsection{Evidence Retriever:}
To resolve relational conflicts, the system first identifies the involved entities and invokes an Evidence Retriever agent to collect supporting or contradicting evidence. This agent adopts a hybrid retrieval strategy that combines both internal and real-time sources. For internal retrieval, each PubMed abstract is encoded using a \textit{all-MiniLM-L6-v2} Sentence Transformer and the query $q_r$, derived from the conflicting relation $r$, is also embedded in the same space, and cosine similarity is computed between the query vector and all document embeddings. The top-$k$ most similar abstracts are retrieved as potential evidence. For real-time evidence, the agent leverages the \texttt{DuckDuckGoSearch} tool from LangChain to issue entity pair-based natural language queries and retrieve relevant scientific updates from the web. Retrieved passages are parsed and filtered using domain-specific keywords, and metadata such as publication source, title, and date are retained. This dual-source retrieval ensures both semantically aligned historical context and up-to-date scientific information are considered for robust relational validation.

\subsubsection{Evidence Evaluator:}
The Evidence Evaluator agent leverages pretrained knowledge and domain-specific priors to assign scores for each evidence passage $e$ based on following three key aspects:

\begin{itemize}
    \item \textbf{Source Reliability (\(r_s\))}: The LLM agent identifies the type of source (e.g., peer-reviewed article, government report, web content) and assigns a credibility score in $[0,1]$, based on its exposure to high-quality sources during pretraining.
    
    \item \textbf{Timeliness (\(r_t\))}: The LLM agent evaluates the recency of the document by computing a normalized freshness score in $[0,1]$, where more recent publications are typically scored higher.
    
    \item \textbf{Relevance (\(r_c\))}: The relevance score $r_c \in [0, 1]$ is computed as a weighted sum of semantic similarity and linguistic certainty: $r_c = \alpha \cdot \text{Sim}(q_r, e) + (1 - \alpha) \cdot \text{CM}_{\text{LLM}}(e)$, where $\text{Sim}(q, e)$ is the cosine similarity between the query and evidence embeddings, and $\text{CM}_{\text{LLM}}(e)$ is a certainty score predicted by LLM agent based on cues like assertive(e.g., "demonstrates", "strongly associated with"), hedging expressions (e.g., "may", "unclear").
\end{itemize}

A final composite score \(s(e)\) is computed as: $s(e) = \lambda_1 r_s + \lambda_2 r_t + \lambda_3 r_c$, where $\lambda_1$, $\lambda_2$, $\lambda_3$ are adaptive weights that reflect the relative importance of each factor. 

\subsubsection{Contradiction Resolver:}
The Contradiction Resolver LLM agent first performs stance classification by labeling each evidence passage as `supporting', `opposing', or `neutral' with respect to the target relation $x \xrightarrow{r} y$. To quantify the aggregate support for the relation, a confidence score $\tau_{x \xrightarrow{r} y}$ is computed as:
\begin{align*}
\tau_{x\xrightarrow{r}y} = 
    &\frac{ \sum\limits_{e \in E_{\text{sup}}(x \xrightarrow{r} y)} s(e) 
    - \sum\limits_{e \in E_{\text{opp}}(x \xrightarrow{r} y)} s(e) }
    { \sum\limits_{e \in E(x \xrightarrow{r} y)} s(e) }
\end{align*}
, where $s(e)$ is the evidence score from the Evidence Evaluator; $E_{\text{sup}}$ and $E_{\text{opp}}$ denote supporting and opposing evidence sets, and $E$ is the full retrieved set.  If $\tau_{x \xrightarrow{r} y} \geq 0.8$, the relation is accepted as valid; otherwise, it is replaced with its negated form $x \xrightarrow{\text{not\_r}} y$. The agent also generates a natural language justification for each decision, and the resulting validated or revised relation is integrated into the Toxicity Trajectory Graph.




\section{Results and Discussion}
\begin{table*}[ht]
\small

\setlength\tabcolsep{2pt}
\centering
\caption{Comparison of relation extraction performance of diffrent baseline models with our Relational Metapath Extraction module \ref{metapath_extraction}, accross different relation types. S: Source, P: pollutant, M: Medium, R: Exposure Route, O: Organ, D: Disease}
\begin{tabular}{|l|lll|lll|lll|lll|lll|}

\hline
Models & \multicolumn{3}{c|}{S $\xrightarrow{\text{emits}}$ P} & \multicolumn{3}{c|}{P $\xrightarrow{\text{contaminates}}$ M} & \multicolumn{3}{c|}{P $\xrightarrow{\text{consumed}}$ R} & \multicolumn{3}{c|}{P $\xrightarrow{\text{affects}}$ O} & \multicolumn{3}{c|}{P $\xrightarrow{\text{causes}}$ D} \\
\hline
 & P& R& F     & P& R & F     & P & R & F     & P & R & F     & P & R & F \\
\hline
BioBERT &  0.56&0.58&0.57	&	0.41&0.47&0.45	&	0.47&0.50&0.48	&	0.42&0.46&0.45	&	0.57&0.55&0.58 \\

\hline
 Llama-3.1-8B-Instruct &	0.61&0.68&0.63	&	0.61&0.59&0.59	&	0.63&0.60&0.62	&	\textbf{0.61}&\textbf{0.65}&\textbf{0.65}	&	0.69&0.68&0.68\\

\hline
 our model \ref{metapath_extraction} &	\textbf{0.73}&\textbf{0.70}&\textbf{0.72}	&	\textbf{0.68}&\textbf{0.62}&\textbf{0.65}	&	\textbf{0.67}&\textbf{0.61}&\textbf{0.65}	&	0.60&0.62&0.61	&	\textbf{0.75}&\textbf{0.74}&\textbf{0.75}\\
\hline
\end{tabular}

\label{results-relation extraction}
\end{table*}

\begin{table*}
\footnotesize
\caption{Summary of top five most frequently identified pollutants in PubMed literature, with their most common sources, affected organs, and associated diseases}
\label{pollutant_summary}
\centering
\renewcommand{\arraystretch}{1}
\begin{tabular}{p{.6in} p{1.3in} p{1.3in}  p{2.3in}} 
\hline
Pollutant & Sources & Affected Organ & Disease Causes\\
\hline
\hline
Polystyrene & plastic products, microbeads, food packaging, infant formula, cosmetics & Gut, Liver, Kidney, Cardiac system, Reproductive system &  metabolic dysfunction, myocardial apoptosis, renal fibrosis, liver damage, spermatogenesis disorder
\\
\hline
Polyethylene & plastic products, food packaging, bottled water, toothpaste, sanitary pads & Gut, Liver, Endocrine system, Vascular system, Hepatopancreas & inflammatory bowel disease, endocrine disruption, liver hyperplasia, hepatodystrophy, necrosis, hypergammaglobulinemia
\\
\hline
Polyvinyl chloride & plastic packaging, water pipes, toys, manufacturing plants waste, IV bags & Liver, Gut, Esophagus, Kidney, Cardiac system & gut microbiota dysbiosis, myocardial infarction, hepatocyte damage, dysphagia , kidney oxidative damage
\\
\hline
Polypropylene & dietary supplements, face masks, plastic wastes, textiles, pp-bottled injections & Gut, Liver, Cardiac system, Kidney, Immune system & metabolism disorder, immune dysfunction, kidney injury, liver dysfunction,  lipid disorder
\\
\hline
Bisphenol A & epoxy resins, dental fillings, infant bottles, food containers, toys & Liver, Gut, Breast, Nervous system, Prefrontal cortex & fatty liver disease, endocrine disruption, obesity, cortical neuritogenesis, infertility
\\
\hline
\end{tabular}
\end{table*}

\begin{figure*}
\centering
\subfloat[]{\includegraphics[width = 3in]{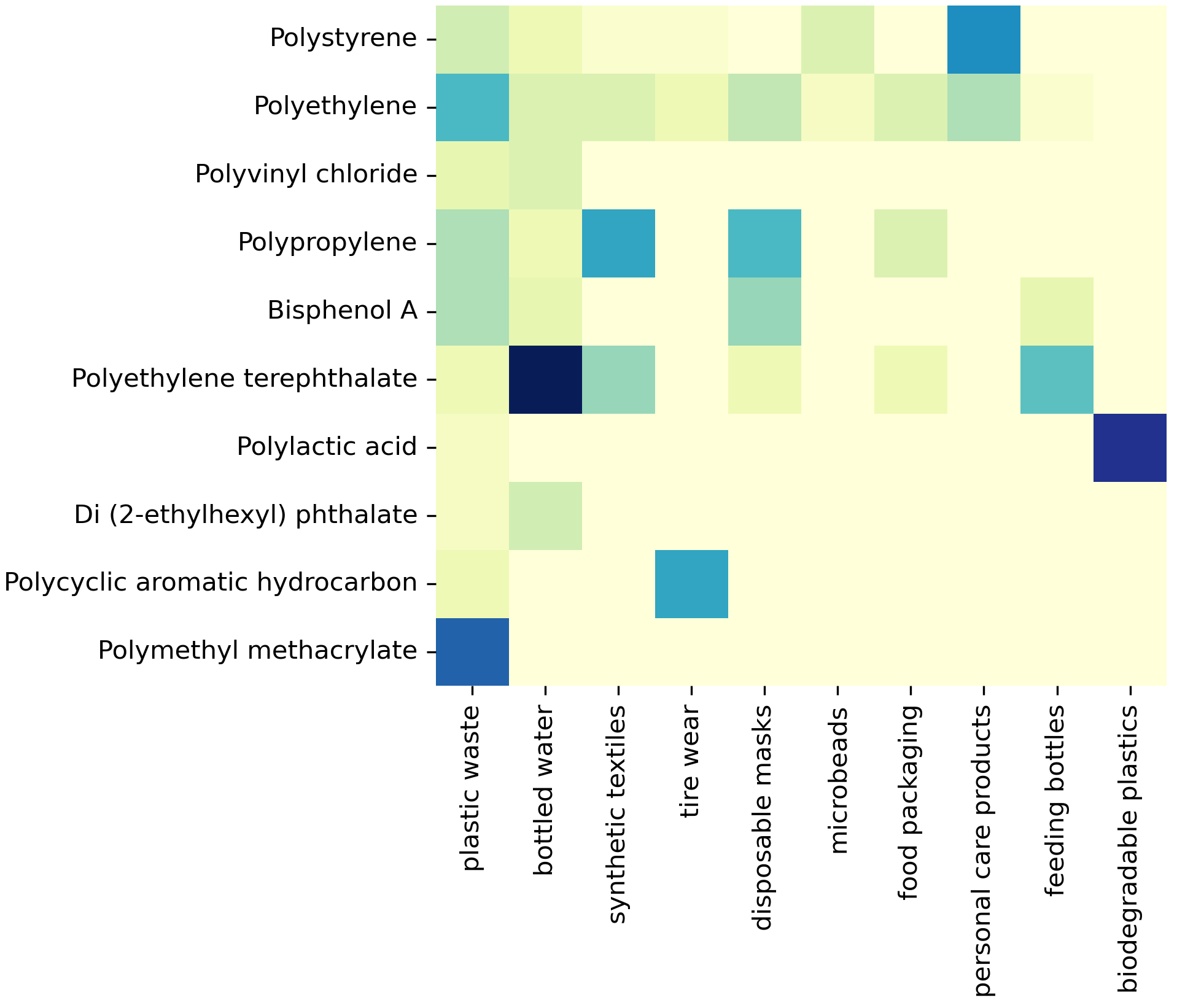}} 
\subfloat[]{\includegraphics[width = 3in]{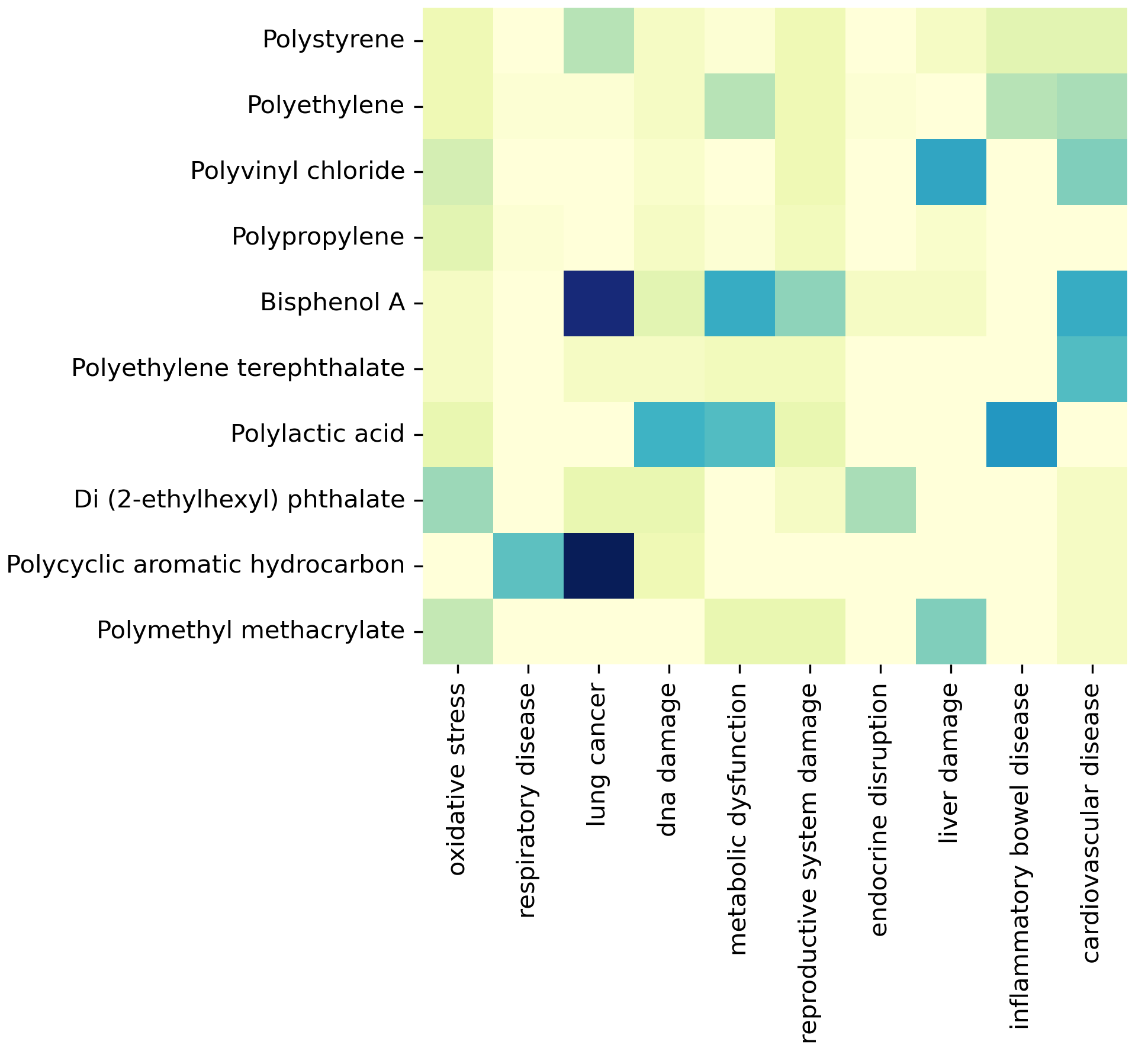}}
\caption{(a)Heatmap showing correlations between most frequent pollutants and most probable sources (b)Heatmap illustrating prevalent causal relationships between these pollutants and associated diseases. Darker colors indicate stronger associations, while lighter colors represent weaker ones.}
\label{heatmap_correlation}
\end{figure*}
We evaluated our framework on 5,282 curated PubMed abstracts, filtered to retain only those describing pollutant-induced diseases or physiological abnormalities. Relational metapaths involving various pollutants were extracted using our Relational Metapath Extraction module (Section~\ref{metapath_extraction}). To assess relation extraction performance, a gold-standard dataset of 1,000 abstracts was manually annotated by domain experts with entity-relation triplets. Evaluation metrics—Precision, Recall, and F1-score—were computed using exact match criteria on both entities and relations. As shown in Table~\ref{results-relation extraction}, our model consistently outperformed baseline models across most relation categories, achieving the highest F1-score and demonstrating its effectiveness in capturing fine-grained, domain-specific semantics.

Furthermore, to ensure consistency in these relations, a relational consistency module detected and resolved conflicting relations by computing confidence scores through a disagreement resolution mechanism. Since no gold labels are available for validating relational conflicts, we conducted an intrinsic evaluation of the disagreement resolution model. We sampled 100 high-confidence resolved relations $(\tau \geq 0.8)$ and found that 92\% were judged correct upon manual inspection by domain experts. Table \ref{confidence_score} in Appendix \ref{conficting_relations_appendix} presents some examples of such conflicting relations. We observed that the majority of these conflicts occurred between pollutants and diseases, with a smaller number involving contradictions between pollutants and their identified sources. Finally, we obtained 49,280 unique relational metapaths encompassing 316 distinct pollutants from 5,282 abstracts.



Table~\ref{pollutant_summary} presents the top five most frequently occurring pollutants in the corpus, along with their dominant sources, affected organs, and associated diseases. While these pollutants are commonly introduced into the environment through food, water, soil, and air, human exposure typically occurs via ingestion, inhalation, dermal absorption, bloodstream entry, or maternal transmission. Given the consistency of these exposure routes across pollutants, they are omitted from the table for clarity.

Finally, we constructed the Toxicity Trajectory Knowledge Graph, summarized in Table \ref{KG_summary} in Appendix \ref{TTG_appendix}, encompassing 316 distinct pollutants originating from 2,134 sources, impacting 297 organs or physiological systems, and associated with 2,772 diseases. Approximately 58\% of abstracts referred to ``microplastic'' or ``nanoplastic'' without specifying exact pollutant names; thus, these terms were explicitly modeled as pollutant nodes to preserve relational coverage. To maintain structural consistency in cases where exposure medium, route, or target organ were unspecified, we introduced `unknown' placeholder nodes. To improve granularity, disease nodes were further enriched with affected population groups, when reported, enabling finer-grained demographic-level analysis.

In a subsequent analysis, we conducted a correlation study between the most frequently occurring pollutants extracted from our dataset, their probable sources, and the diseases they most commonly cause. The heatmaps in Figure \ref{heatmap_correlation} provide comprehensive visualizations of the pollutant-source correlations (Figure a) and prevalent pollutant-disease causal relationships (Figure b). In these heatmaps, darker colors represent stronger correlations, while lighter colors indicate weaker ones. We observed that \textit{bottled water} is a significant source of \textit{polyethylene terephthalate} exposure through water intake, with \textit{feeding bottles} also being a major source for neonates, has been linked to cardiovascular diseases. Studies have identified \textit{disposable masks} as a source of pollutants such as \textit{polypropylene} and \textit{Bisphenol A}, which are associated with diseases including \textit{lung cancer}, \textit{metabolic dysfunction}, and \textit{cardiovascular diseases}. \textit{Polycyclic aromatic hydrocarbons}, originating from \textit{tire wear}, are another major pollutant, primarily linked to \textit{lung cancer}. Although \textit{biodegradable plastics} are often regarded as environmentally friendly, they release \textit{Polylactic acid}, according to some studies, which has been associated with several diseases such as \textit{inflammatory bowel disease, DNA damage, and metabolic dysfunction}.

\begin{figure}[ht]
\includegraphics[width=\linewidth]{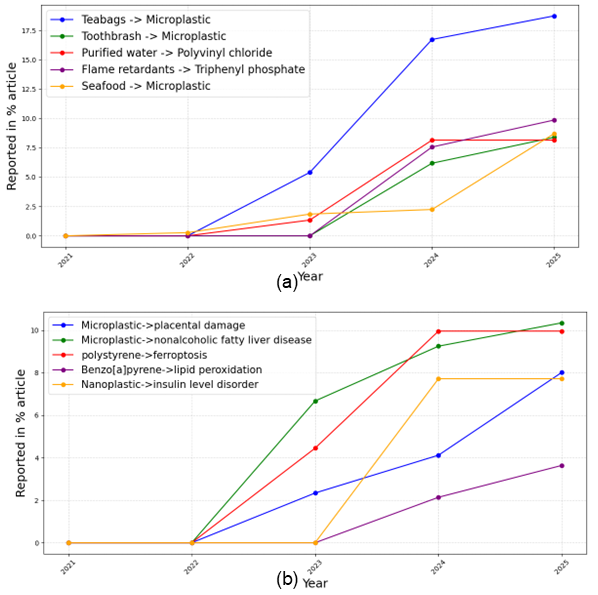}
\centering
\caption{Illustrating trends in newly reported relations over time : (a) Emerging sources of plastic pollutants and (b) diseases associated with pollutants.}
\label{time_analysis}
\end{figure}

Moreover, we conducted a longitudinal analysis to assess the emergence of new pollutants, their sources, and associated diseases related to plastic pollutants in recent years. Given the evolving nature of this field, it is crucial to monitor new investigations and findings to inform preventive measures for public health. Figure \ref{time_analysis} highlights some examples of emerging pollutant sources and associated diseases identified in recent PubMed abstracts. Several recent studies have reported microplastic contamination originating from everyday items such as \textit{teabags, toothbrushes}, and \textit{seafood}, with ingestion linked to various \textit{gastrointestinal disorders}. Additionally, the presence of \textit{PVC nanoparticles} has been detected in \textit{purified water}, raising concerns about their potential health risks.

Emerging evidences also suggest microplastics exposure adversely affect the reproductive system disorder including \textit{placental damage} and \textit{impaired spermatogenesis}, as well as in broader physiological conditions such as \textit{insulin resistance}, \textit{fatty liver diseases} and so on. These findings underscore the need for continuous surveillance and the promotion of safer material alternatives to mitigate health risks associated with plastic toxicity.

\section{Conclusion}
In this study, we presented an LLM-based framework for extracting relational metapaths and finally constructing a toxicity trajectory knowledge graph of plastic pollutants based on large-scale analysis of PubMed abstracts.
The resulting knowledge graph uncover both direct and complex interrelations among pollutants, their sources, exposure pathways, and associated health outcomes, providing a valuable resource for advancing the understanding of plastic toxicity and its public health implications. In future work, we plan to integrate multimodal data to investigate the causal effects of pollutant exposure, accounting for confounding factors such as geographic variability, socioeconomic status, and lifestyle behaviors.

\section*{Limitations}
While Our proposed framework effectively extracts relational metapaths and constructs a comprehensive toxicity trajectory graph, it exhibits certain limitations. The reliance on abstracts rather than full-text articles may lead to incomplete or context-deficient relation extraction. The absence of gold-standard annotations for contradiction resolution necessitates heuristic-based evaluation, which may introduce subjective bias. We are addressing these limitations by incorporating full-text analysis, and exploring semi-supervised validation and contextual disambiguation to improve relational accuracy and demographic specificity.


\bibliography{custom}

\begin{thebibliography}{24}
\providecommand{\natexlab}[1]{#1}

\bibitem[{Al~Mamun et~al.(2023)Al~Mamun, Prasetya, Dewi, and Ahmad}]{al2023microplastics}
Abdullah Al~Mamun, Tofan Agung~Eka Prasetya, Indiah~Ratna Dewi, and Monsur Ahmad. 2023.
\newblock Microplastics in human food chains: Food becoming a threat to health safety.
\newblock \emph{Science of the Total Environment}, 858:159834.

\bibitem[{Amato-Louren{\c{c}}o et~al.(2020)Amato-Louren{\c{c}}o, dos Santos~Galv{\~a}o, de~Weger, Hiemstra, Vijver, and Mauad}]{amato2020emerging}
Lu{\'\i}s~Fernando Amato-Louren{\c{c}}o, Luciana dos Santos~Galv{\~a}o, Letty~A de~Weger, Pieter~S Hiemstra, Martina~G Vijver, and Thais Mauad. 2020.
\newblock An emerging class of air pollutants: potential effects of microplastics to respiratory human health?
\newblock \emph{Science of the total environment}, 749:141676.

\bibitem[{Aristizabal et~al.(2024)Aristizabal, Jim{\'e}nez-Orrego, Caicedo-Le{\'o}n, P{\'a}ez-C{\'a}rdenas, Castellanos-Garc{\'\i}a, Villalba-Moreno, Ram{\'\i}rez-Zuluaga, Hsu, Jaller, and Gold}]{aristizabal2024microplastics}
Miguel Aristizabal, Katherine~V Jim{\'e}nez-Orrego, Mar{\'\i}a~D Caicedo-Le{\'o}n, Laura~S P{\'a}ez-C{\'a}rdenas, Isabella Castellanos-Garc{\'\i}a, Dennys~L Villalba-Moreno, Luisa~V Ram{\'\i}rez-Zuluaga, Jeffrey~TS Hsu, Jose Jaller, and Michael Gold. 2024.
\newblock Microplastics in dermatology: Potential effects on skin homeostasis.
\newblock \emph{Journal of Cosmetic Dermatology}, 23(3):766--772.

\bibitem[{Bl{\"o}cker et~al.(2020)Bl{\"o}cker, Watson, and Wichern}]{blocker2020living}
Lisa Bl{\"o}cker, Conor Watson, and Florian Wichern. 2020.
\newblock Living in the plastic age-different short-term microbial response to microplastics addition to arable soils with contrasting soil organic matter content and farm management legacy.
\newblock \emph{Environmental Pollution}, 267:115468.

\bibitem[{Bodenreider(2004)}]{bodenreider2004unified}
Olivier Bodenreider. 2004.
\newblock The unified medical language system (umls): integrating biomedical terminology.
\newblock \emph{Nucleic acids research}, 32(suppl\_1):D267--D270.

\bibitem[{Cai et~al.(2021)Cai, Xu, Du, Li, Liu, and Shi}]{cai2021analysis}
Huiwen Cai, Elvis~Genbo Xu, Fangni Du, Ruilong Li, Jingfu Liu, and Huahong Shi. 2021.
\newblock Analysis of environmental nanoplastics: Progress and challenges.
\newblock \emph{Chemical Engineering Journal}, 410:128208.

\bibitem[{Cverenk{\'a}rov{\'a} et~al.(2021)Cverenk{\'a}rov{\'a}, Valachovi{\v{c}}ov{\'a}, Mackul'ak, {\v{Z}}emli{\v{c}}ka, and B{\'\i}ro{\v{s}}ov{\'a}}]{cverenkarova2021microplastics}
Kl{\'a}ra Cverenk{\'a}rov{\'a}, Martina Valachovi{\v{c}}ov{\'a}, Tom{\'a}{\v{s}} Mackul'ak, Luk{\'a}{\v{s}} {\v{Z}}emli{\v{c}}ka, and Lucia B{\'\i}ro{\v{s}}ov{\'a}. 2021.
\newblock Microplastics in the food chain.
\newblock \emph{Life}, 11(12):1349.

\bibitem[{De-la Torre(2020)}]{de2020microplastics}
Gabriel~Enrique De-la Torre. 2020.
\newblock Microplastics: an emerging threat to food security and human health.
\newblock \emph{Journal of food science and technology}, 57(5):1601--1608.

\bibitem[{Dokl et~al.(2024)Dokl, Copot, Krajnc, Van~Fan, Vujanovi{\'c}, Aviso, Tan, Kravanja, and {\v{C}}u{\v{c}}ek}]{dokl2024global}
Monika Dokl, Anja Copot, Damjan Krajnc, Yee Van~Fan, Annamaria Vujanovi{\'c}, Kathleen~B Aviso, Raymond~R Tan, Zdravko Kravanja, and Lidija {\v{C}}u{\v{c}}ek. 2024.
\newblock Global projections of plastic use, end-of-life fate and potential changes in consumption, reduction, recycling and replacement with bioplastics to 2050.
\newblock \emph{Sustainable Production and Consumption}, 51:498--518.

\bibitem[{Gasperi et~al.(2018)Gasperi, Wright, Dris, Collard, Mandin, Guerrouache, Langlois, Kelly, and Tassin}]{gasperi2018microplastics}
Johnny Gasperi, Stephanie~L Wright, Rachid Dris, France Collard, Corinne Mandin, Mohamed Guerrouache, Val{\'e}rie Langlois, Frank~J Kelly, and Bruno Tassin. 2018.
\newblock Microplastics in air: are we breathing it in?
\newblock \emph{Current Opinion in Environmental Science \& Health}, 1:1--5.

\bibitem[{Huang et~al.(2022)Huang, Huang, Bi, Guo, Yu, Zeng, Huang, Liu, Wu, Chen et~al.}]{huang2022detection}
Shumin Huang, Xiaoxin Huang, Ran Bi, Qiuxia Guo, Xiaolin Yu, Qinghui Zeng, Ziyu Huang, Tianming Liu, Haisheng Wu, Yuliang Chen, and 1 others. 2022.
\newblock Detection and analysis of microplastics in human sputum.
\newblock \emph{Environmental science \& technology}, 56(4):2476--2486.

\bibitem[{Jiang et~al.(2022)Jiang, Han, Na, Fang, Qi, Lu, Liu, Zhou, Feng, Zhu et~al.}]{jiang2022exposure}
Ying Jiang, Jinchi Han, Jun Na, Jing Fang, Chanchan Qi, Junge Lu, Xiaojing Liu, Changhe Zhou, Jing Feng, Weiwei Zhu, and 1 others. 2022.
\newblock Exposure to microplastics in the upper respiratory tract of indoor and outdoor workers.
\newblock \emph{Chemosphere}, 307:136067.

\bibitem[{Lu et~al.(2023)Lu, Li, Wang, Tu, Qiu, Zhang, Zhong, Li, Liu, Liu et~al.}]{lu2023new}
Wenfeng Lu, Xiaoliang Li, Shuguang Wang, Changli Tu, Lan Qiu, Han Zhang, Chenghui Zhong, Saifeng Li, Yuewei Liu, Jing Liu, and 1 others. 2023.
\newblock New evidence of microplastics in the lower respiratory tract: inhalation through smoking.
\newblock \emph{Environmental Science \& Technology}, 57(23):8496--8505.

\bibitem[{MacLeod et~al.(2021)MacLeod, Arp, Tekman, and Jahnke}]{macleod2021global}
Matthew MacLeod, Hans Peter~H Arp, Mine~B Tekman, and Annika Jahnke. 2021.
\newblock The global threat from plastic pollution.
\newblock \emph{Science}, 373(6550):61--65.

\bibitem[{Menichetti et~al.(2024)Menichetti, Mordini, and Montalti}]{menichetti2024penetration}
Arianna Menichetti, Dario Mordini, and Marco Montalti. 2024.
\newblock Penetration of microplastics and nanoparticles through skin: Effects of size, shape, and surface chemistry.
\newblock \emph{Journal of Xenobiotics}, 15(1):6.

\bibitem[{Ouyang et~al.(2025)Ouyang, Wu, Zhao, Hu, Jiang, Fu, Lei, Zhang, Duan, Huang et~al.}]{ouyang2025microplastics}
Yujie Ouyang, Songjiang Wu, Yuanyuan Zhao, Yibo Hu, Ling Jiang, Chuhan Fu, Li~Lei, Yushan Zhang, Xiaolei Duan, Jinhua Huang, and 1 others. 2025.
\newblock Microplastics and skin aging: Disruption of barrier function and induction of fibroblast senescence.
\newblock \emph{Experimental Dermatology}, 34(1):e70027.

\bibitem[{Rainieri and Barranco(2019)}]{rainieri2019microplastics}
Sandra Rainieri and Alejandro Barranco. 2019.
\newblock Microplastics, a food safety issue?
\newblock \emph{Trends in food science \& technology}, 84:55--57.

\bibitem[{Reimers and Gurevych(2019)}]{reimers-2019-sentence-bert}
Nils Reimers and Iryna Gurevych. 2019.
\newblock \href {https://arxiv.org/abs/1908.10084} {Sentence-bert: Sentence embeddings using siamese bert-networks}.
\newblock In \emph{Proceedings of the 2019 Conference on Empirical Methods in Natural Language Processing}. Association for Computational Linguistics.

\bibitem[{Riaz et~al.(2024)Riaz, Lodhi, Munir, Zhao, Farooq, Qadri, and Islam}]{riaz2024breathing}
Hafiz~Hamza Riaz, Abdul~Haseeb Lodhi, Adnan Munir, Ming Zhao, Umar Farooq, M~Qadri, and Mohammad~S Islam. 2024.
\newblock Breathing in danger: mapping microplastic migration in the human respiratory system.
\newblock \emph{Physics of Fluids}, 36(4).

\bibitem[{Ritchie et~al.(2023)Ritchie, Samborska, and Roser}]{owid-plastic-pollution}
Hannah Ritchie, Veronika Samborska, and Max Roser. 2023.
\newblock Plastic pollution.
\newblock \emph{Our World in Data}.
\newblock Https://ourworldindata.org/plastic-pollution.

\bibitem[{Then et~al.(2014)Then, Kaufmann, Chirigati, Hoang-Vu, Pham, Kemper, Neumann, and Vo}]{then2014more}
Manuel Then, Moritz Kaufmann, Fernando Chirigati, Tuan-Anh Hoang-Vu, Kien Pham, Alfons Kemper, Thomas Neumann, and Huy~T Vo. 2014.
\newblock The more the merrier: Efficient multi-source graph traversal.
\newblock \emph{Proceedings of the VLDB Endowment}, 8(4):449--460.

\bibitem[{Thompson et~al.(2004)Thompson, Olsen, Mitchell, Davis, Rowland, John, McGonigle, and Russell}]{thompson2004lost}
Richard~C Thompson, Ylva Olsen, Richard~P Mitchell, Anthony Davis, Steven~J Rowland, Anthony~WG John, Daniel McGonigle, and Andrea~E Russell. 2004.
\newblock Lost at sea: where is all the plastic?
\newblock \emph{Science}, 304(5672):838--838.

\bibitem[{Veli{\v{c}}kovi{\'c} et~al.(2017)Veli{\v{c}}kovi{\'c}, Cucurull, Casanova, Romero, Lio, and Bengio}]{velivckovic2017graph}
Petar Veli{\v{c}}kovi{\'c}, Guillem Cucurull, Arantxa Casanova, Adriana Romero, Pietro Lio, and Yoshua Bengio. 2017.
\newblock Graph attention networks.
\newblock \emph{arXiv preprint arXiv:1710.10903}.

\bibitem[{Wang et~al.(2023)Wang, Xu, and Jiang}]{wang2023microplastics}
Yuchen Wang, Xinqi Xu, and Guan Jiang. 2023.
\newblock Microplastics exposure promotes the proliferation of skin cancer cells but inhibits the growth of normal skin cells by regulating the inflammatory process.
\newblock \emph{Ecotoxicology and Environmental Safety}, 267:115636.

\end{thebibliography}

\appendix
\section{PubMed abstract with highlighted entities}
\label{appendix: sample_abstract}
\begin{figure}[ht]
\includegraphics[width=\linewidth]{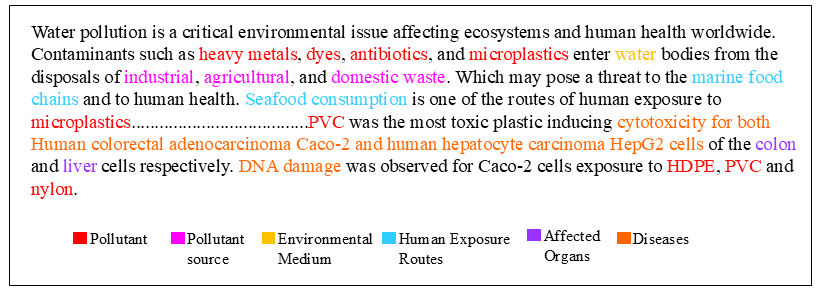}
\centering
\caption{Sample PubMed abstract with highlighted entities.}
\label{sample_abstract}
\end{figure}

\section{User-Defined Queries for Extraction of Relational Metapaths}
\label{User_Defined_Queries}
\begin{tcolorbox}[colback=gray!5!white, colframe=gray!60!black, title = {User Defined Queries (Q) }]
Q1. What pollutants or contaminants are mentioned in the abstract, and what are their respective sources?\\
Q2. Which environmental media (e.g., air, water, soil) are contaminated by these pollutants?\\
Q3. Through which exposure routes (e.g., inhalation, ingestion, dermal contact) do these pollutants reach or expose humans?\\
Q4. Which human organs or biological systems are affected by exposure to these pollutants?\\
Q5. What diseases, health effects, or physiological abnormalities are caused in these organs?\\
Q6. Are specific demographic groups (age group, gender) more affected by these pollutants or health outcomes? If so, specify the group.\\
Q7. Is there any indication that a specific source does not emit a particular pollutant? If so, specify the source and pollutant.\\
Q8. Is there any indication that a particular pollutant does not affect a specific organ? If so, specify the pollutant and organ.\\
Q9. Is there any indication that a particular pollutant is not linked to a specific disease or health effect? If so, specify the pollutant and disease.\\
\label{questions}
\end{tcolorbox}

\section{Example of relational metapath converted into cypher query}
\label{appendix_cypherquery}

\begin{figure*}
\centering
\subfloat[]{\includegraphics[width = 2.5in]{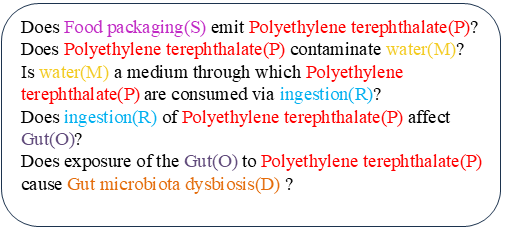}}
\subfloat[]{\includegraphics[width = 2.5in]{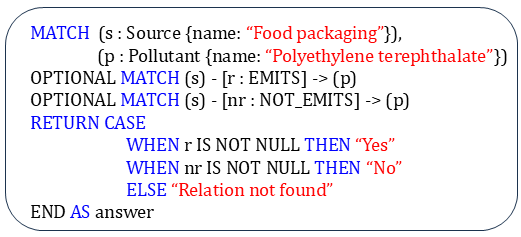}}
\caption{(a) Generated query of a metapath in Fig. \ref{sample_path}. (b) Representation of Cypher query of the query ``Does Food packaging emits Polyethylene terephthalate?''.}
\label{consistency_detection}
\end{figure*}

Consider a metapath shown in Figure \ref{sample_path},
\textit{Food packaging} $\xrightarrow{\text{emits}}$ \textit{Polyethylene terephthalate} $\xrightarrow{\text{contaminates}}$ \textit{water} $\xrightarrow{\text{consumed through}}$ \textit{ingestion}$ \xrightarrow{\text{affects}}$ \textit{Gut} $\xrightarrow{\text{causes}}$ \textit{Gut microbiota dysbiosis}. Figure \ref{consistency_detection}(a) presents set of queries generated for this metapath and figure \ref{consistency_detection} (b) is the representation of cypher query of a query.

\section{Table \ref{confidence_score} contains examples of conflicting relations}
\label{conficting_relations_appendix}
\begin{table*}
\footnotesize
\caption{Examples of some conflicting relations with their confidence score.}
\label{confidence_score}
\centering
\renewcommand{\arraystretch}{1}
\begin{tabular}{p{2in} p{.7in} p{2.3in}} 
\hline
Conflicting Relations & Confidence score & Explanation of Contradiction Resolver\\
\hline
Bisphenol A $->$ women infertility & 0.81 & Evidences suggest that BPA is an endocrine-disrupting chemical that can interfere with reproductive hormones like estrogen, though establishing direct causation in human infertility remains a subject of ongoing investigation \\
\hline
PVC $->$ Liver Cancer & 0.23 & evidences suggest while vinyl chloride, used in PVC production, is a known liver carcinogen, the risk of liver cancer from the finished PVC product under normal use is considered low.\\
\hline
Microplastics $->$ gastrointestinal microbiota dysbiosis & 0.45 & evidences are primarily from animal studies, indicates that microplastic exposure cause gastrointestinal microbiota dysbiosis, very limited human studies provide suggestive links between microplastic and alterations in the gut microbiota, particularly in the context of IBD.\\
\hline
Plastic cutting board $->$ microplastic & 0.85 & evidences strongly suggest that plastic cutting boards are source of microplastic pollution in food.\\
\hline

\end{tabular}

\end{table*}

\section{Table \ref{KG_summary} summarizes Toxicity Trajectory Graph}
\label{TTG_appendix}
\begin{table*}
\caption{Summary of Toxicity Trajectory Graph.}
\label{KG_summary}
\centering
\renewcommand{\arraystretch}{1.5}
\begin{tabular}{p{2in} p{3in}} 
\hline
Characteristics & value\\
\hline
Node types & Source, Pollutant, Environmental medium, Exposure route, Affected organ, Disease.\\
\hline
No. of nodes & Source: 2,134; Pollutant :316 Medium : 5; Exposure route: 6; Affected organ 297; Disease : 2,772\\
\hline
Edge types & Emit, Contaminate, Consume through, Affect, Cause, not\_emit, not\_cause\\
\hline
No. of edges & 21,384\\
\hline

\end{tabular}

\end{table*}


\end{document}